# Automating Manual Tasks through Intuitive Robot Programming and Cognitive Robotics

Bijan KAVOUSIAN, Petar TESIC, Oliver PETROVIC, Christian BRECHER

*Laboratory for Machine Tools and Production Engineering (WZL)
of RWTH Aachen University,
Steinbachstraße 19, D-52074 Aachen*

**Abstract:** This paper presents a novel concept for intuitive end-user programming of robots, inspired by natural interaction between humans. Natural language and supportive gestures are translated into robot programs using large language models (LLMs) and computer vision (CV). Through equally natural system feedback in the form of clarification questions and visual representations, the generated program can be reviewed and adjusted, thereby ensuring safety, transparency, and user acceptance.

**Key Words:** End-User Programming, Robotics, User Interface, Large Language Models, Human-Machine-Interaction

## 1   Introduction

The automation of manual tasks in industrial manufacturing is increasingly becoming an economic necessity. Monotonous and ergonomically demanding tasks are perceived as unpleasant by workers, while companies simultaneously face rising labor costs (Statistisches Bundesamt, 2024; Thackray, 1981). According to a survey by Knapp et al. (2022), 80% of the surveyed companies experience significant cost pressure, with automation frequently cited as a mitigation measure.

The increasing customization of products and the associated reduction in batch sizes impose higher demands on the flexibility and efficiency of automated systems (Schuh et al., 2020). Especially for small batch sizes, setup costs become a decisive factor. In addition, the growing shortage of skilled workers presents a major challenge. Many companies struggle to find qualified personnel, including experts required for robot programming (IHK Frankfurt am Main, 2024). This creates a strong need for intuitive programming methods that are both user-friendly and time-efficient, enabling rapid adaptation of robots to changing production requirements.

## 2   Related Work

Traditional robot programming methods are associated with high effort, require expert knowledge, and are often inflexible. Programs are typically created using teach-in methods, where the robot is manually guided to specific positions that are then stored (Heimann & Guhl, 2020). Changes in processes or products often require substantial reprogramming (Hoebert et al., 2023).

Collaborative robots (cobots) represent a step toward more intuitive programming. Manufacturers provide user interfaces that allow users without deep programming knowledge to define robot workflows (Schmidbauer et al., 2020). This lowers the entry barrier to robotics, however, even minor changes in products or processes still require significant manual adjustments.

Recent advances in artificial intelligence (AI) have also influenced robotics. Systems such as RT-2 enable robots to perform a wide range of manipulation tasks, including previously unseen ones, by directly generating actions from AI models (Brohan et al., 2023). Other approaches use large language models (LLMs) to generate executable robot code (Beschi et al., 2019; Karli et al., 2024; Wang et al., 2024). However, their industrial applicability remains limited due to safety concerns and the unpredictability of AI behavior. Furthermore, purely language-based task descriptions are insufficient for representing spatial relationships and selecting specific objects.

## 3    Requirements for a Robot Programming System

To be applicable in industrial environments, the system must fulfill a number of requirements. These are described below.

**Fast.** In order to be used efficiently, the setup time must be significantly shorter than the time required for manual execution, even for small batch sizes. A fast setup not only reduces direct production costs but also lowers the barrier to using the system for smaller batch sizes or frequently changing tasks. As a result, the system becomes more versatile and attractive for a wider range of applications.

**Intuitive to use.** The system should be usable by workers without in-depth IT or robotics knowledge. This enables process-experienced employees to directly incorporate their domain-specific knowledge into the automation system, without relying on external specialists for commissioning.

**Transparent.** Transparent communication of the system's behavior can increase trust in the system (Kizilcec, 2016; Schraagen et al., 2021). As trust in AI systems increases, so does their acceptance (Choung et al., 2023). Transparency is therefore intended to promote acceptance of the system.

**Human-in-the-loop control.** Despite automation and AI support, humans should always retain control over the system (Sartori & Theodorou, 2022). This includes the ability to intervene, as well as mechanisms that ensure that system decisions can be reviewed and adjusted. Such human-centered control is essential to ensure trust in the system and to meet safety requirements.

**Flexible.** The system must be able to adapt dynamically to changing production conditions. This includes the ability to respond to variations in the position and shape of objects. In addition, the system should be compatible with different robot models and their specific requirements in order to ensure broad applicability.

**Multimodal.** For natural and context-appropriate interaction, different modalities such as language, gestures, and visual cues should be combined. This allows users to go beyond purely text-based commands and instead incorporate their environment dynamically when defining the desired workflow.

## 4   Interaction Concept

Based on these requirements, a concept for intuitive robot programming was developed, which aims to reflect the natural explanation process between two workers (Figure 1). In a typical training situation, interaction takes place through descriptions, gestures, clarification questions, and confirmations in order to define and verify the workflow. This dynamic is intended to be realized through the implementation of multiple modules.

The central element of the programming process is communication via natural language. Users formulate instructions or task goals, which are interpreted by the system and translated into robot actions. Through targeted clarification questions, the system iteratively completes missing parts of the workflow and resolves misunderstandings. By accessing internal databases, models, operating manuals, and bills of materials, the system is further enabled to propose optimizations of the program, which can then be evaluated by the worker.

In addition to language, gestures are often used to convey and describe processes, as they are intuitive and provide additional information. This additional information is also intended to be utilized during program creation, allowing users to point at components, tools, or fixtures, specify positions, or define trajectories. Combined with language, these objects and positions can be enriched with semantic descriptions, which further supports the system in understanding the intended task.

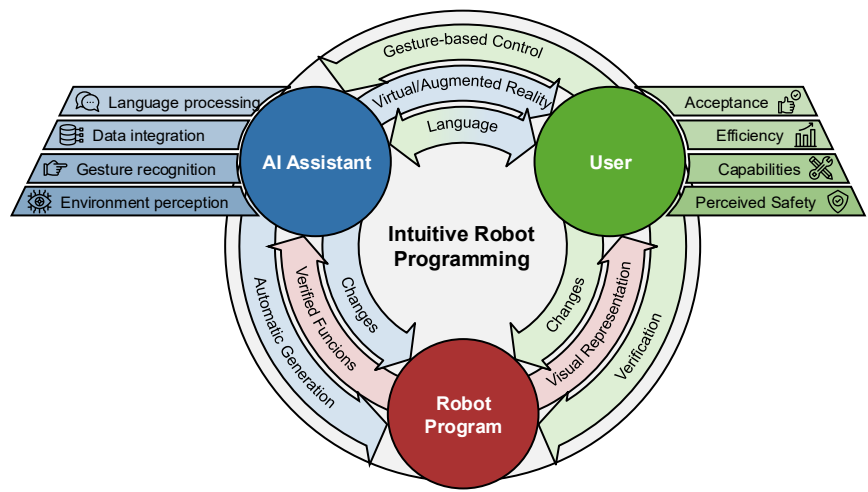

*Figure 1:*   *Concept of intuitive robot programming based on multimodal interaction between user and AI assistant, integrating language, gestures, and augmented reality to iteratively generate, verify, and refine robot programs.*

Beyond the dialog, visual feedback should also be provided to the users. A central challenge is to represent positions and trajectories in space in a way that is understandable for the user. To facilitate this, an augmented reality (AR) application is integrated, which enhances the current scene with additional digital information. This visualization enables users to directly see and understand which positions and motion trajectories have been defined. At the same time, the corresponding information derived from the verbal description is displayed in a visually linked manner. Through this enhanced visualization, users can not only better understand the program flow but also edit and adjust it more efficiently. This increases the transparency of the robot control and significantly simplifies the validation of the programmed processes, thereby further improving the usability of the system.

With the validated programs, the robot can execute tasks repeatedly and without constant supervision. In addition, previously validated functions can be reused in new programs. This reduces the need to write new code and minimizes the potential for errors.

## 5   Technical Concept

The technical design of the described interaction concept is based on a combination of modern AI and AR technologies. As a central component for processing natural language and generating the robot program, a large language model (LLM) is employed. Such models are capable of understanding complex, multi-step tasks within their context and structuring them accordingly. This includes identifying subtasks, logically sequencing work steps, and generating coherent programs tailored to the target requirements (Zhou et al., 2024). Furthermore, LLMs can translate natural language into standardized data formats, enabling further processing by robotic systems. For this purpose, the LLM is provided with documentation of the available basic functions, explaining the purpose and functionality of each building block. To ensure reusability of automatically generated functions, documentation is also created and stored alongside each function. This allows the LLM to reuse previously validated functional blocks.

Modern VR and AR headsets provide the necessary hardware for hand and gesture interaction as well as for three-dimensional visualization in space. New waypoints can, for example, be defined by tapping in mid-air and are directly visualized. Through a short verbal description provided by the user, the LLM can enrich these points with additional attributes, assigning them specific functions that are reflected in the visualization.

For a basic understanding of the working environment, advanced computer vision (CV) models can be utilized. Segment Anything Models (SAM) are capable of segmenting and identifying entities within an image (Kirillov et al., 2023). In combination with gesture recognition, this enables the selection of objects within the environmental context and their additional semantic description.

## 6   Discussion

The presented concept addresses key challenges of modern production environments, such as increasing demands for flexible production processes and the associated complexity of commissioning procedures. LLMs, advanced computer vision (CV) approaches, and multimodal interfaces offer the possibility to integrate natural language, gestures, and other context-related inputs into robot control. These technologies provide the foundation for intuitive and seamless human-machine interaction, with the potential to fundamentally transform automation. Despite these advances, it remains an open research question which specific types of information and multimodal inputs are actually required for an interaction to be perceived as "natural" and to adequately represent the complexity and dynamics of real-world work processes. Future investigations will particularly focus on ergonomics, effectiveness, and user acceptance of such systems. As a next step, a demonstrator based on the presented concept will be developed to evaluate the system across various use cases.

Using the developed system, further research will investigate how natural interaction modalities can contribute to making the automation of manual tasks more accessible and efficient. However, practical application will ultimately depend on the usability and robustness of the developed solutions in meeting the requirements of industrial environments. Successful integration requires close alignment with user needs as well as the continuous development of human-robot interaction models.

**Acknowledgements:** The studies described in this paper were conducted as part of the research and development project "AKzentE4.0". The project is funded by the German Federal Ministry of Research, Technology and Space (BMFTR) under the funding measure "Future of Work: Regional Competence Centers of Labor Research. Designing New Forms of Work through Artificial Intelligence" within the program "Innovations for Tomorrow's Production, Services and Work" (funding code: 02L19C400) and supervised by the Project Management Agency Karlsruhe (PTKA).

# Automatisierung manueller Tätigkeiten durch intuitive Programmierung und kognitive Robotik

Bijan KAVOUSIAN, Petar TESIC, Oliver PETROVIC, Christian BRECHER

*Werkzeugmaschinenlabor WZL der RWTH Aachen,
Steinbachstraße 19, D-52074 Aachen*

**Kurzfassung:** In diesem Beitrag wird ein neuartiges Konzept zur intuitiven Endnutzerprogrammierung von Robotern vorgestellt, welches sich an der natürlichen Interaktion zwischen zwei Menschen orientiert. Natürliche Sprache und unterstützende Gestik werden durch den Einsatz von Large Language Models (LLM) und Computer Vision (CV) in Roboterprogramme überführt. Durch ebenso natürliche Rückfragen des Systems und visueller Darstellung kann das Programm überprüft und angepasst werden, wodurch die Sicherheit, Transparenz und Akzeptanz des Systems sichergestellt werden.

**Schlüsselwörter:** Endnutzerprogrammierung, Robotik, User Interface, Large Language Models, Mensch-Maschine-Interaktion

## 1. Einleitung/Motivation

Die Automatisierung manueller Tätigkeiten in der industriellen Fertigung wird zunehmend zu einer wirtschaftlichen Notwendigkeit. Monotone und ergonomisch belastende Tätigkeiten werden von Mitarbeitenden als unangenehm empfunden, während Unternehmen gleichzeitig mit steigenden Lohnkosten konfrontiert sind (Statistisches Bundesamt 2024; Thackray 1981). Laut einer Umfrage von Knapp et al. (2022) verspüren 80 % der befragten Unternehmen einen erheblichen Kostendruck, wobei Automatisierung häufig als Entlastungsmaßnahme genannt wird.

Die zunehmende Individualisierung von Produkten und die damit einhergehende Verringerung der Losgrößen stellen höhere Anforderungen an die Flexibilität und Effizienz automatisierter Systeme (Schuh et al. 2020). Besonders bei geringen Stückzahlen werden Rüstkosten zu einem entscheidenden Faktor. Zusätzlich bedeutet der wachsende Fachkräftemangel eine Herausforderung: Laut der IHK Frankfurt am Main (2024) haben viele Unternehmen Schwierigkeiten, qualifiziertes Personal zu finden. Diese Engpässe betreffen auch spezialisierte Aufgaben wie die Programmierung von Robotern, für die oft spezielles Fachwissen erforderlich ist. Daher entsteht ein dringender Bedarf an intuitiven Programmiermethoden, die sowohl benutzerfreundlich als auch zeiteffizient sind und eine schnelle Anpassung von Robotern an sich wandelnde Produktionsanforderungen ermöglichen.

## 2. Stand der Technik

Klassische Programmiermethoden von Robotern bedeuten einen hohen Aufwand, erfordern Expertenwissen und sind häufig wenig flexibel. Programme werden dabei häufig mit der Teach-In Methode erstelle, bei der mit dem Roboter nacheinander spezifische Positionen angefahren und gespeichert werden (Heimann & Guhl 2020). Ände-rungen in den Prozessabläufen oder den Produkten erfordern oft eine umfangreiche Anpassung oder Neuerstellung der Programme. (Hoebert et al. 2023) Einen Schritt hin zur intuitiven Programmierung von Robotern gehen Cobots, die häufig intuitiv eingerichtet (geteacht) werden können. Dafür stellen die Roboterhersteller Benutzeroberflächen bereit, über die auch ohne tiefe Programmierkenntnisse Abläufe für den Roboter definiert werden können. (Schmidbauer et al. 2020). Dies senkt zwar die Einstiegshürde, jedoch erfordern selbst geringe Änderungen am Produkt oder Prozess umfangreiche manuelle Anpassungen des Programms.

Die aktuellen Fortschritte im Bereich der künstlichen Intelligenz (KI) treiben auch die Entwicklung in der Robotik weiter voran. So wurde mit RT2 ein System entwickelt, das eine Vielzahl auch unbekannter Manipulationsaufgaben ausführen kann, indem die KI dem Roboter direkt eine Bewegung vorgibt (Brohan et al. 2023). Andere Ansätze verwenden Large Language Models (LLMs), um Programmcode zu generieren, der vom Roboter ausgeführt wird (Beschi et al. 2019; Karli et al. 2024; Wang et al. 2024). Dennoch bleibt die wirtschaftliche Anwendbarkeit dieser Ansätze bislang begrenzt, da die Sicherheitsanforderungen der Anwendungen nur schwer mit dem unvorhersehbaren Verhalten der KI vereinbar sind. Zudem basiert die Aufgabenbeschreibung ausschließlich auf Sprache, was insbesondere bei der Darstellung räumlicher Beziehungen und der Auswahl spezifischer Objekte zu Einschränkungen führen kann.

## 3. Anforderungen an ein System zur Programmierung von Robotern

Zur Anwendbarkeit im industriellen Umfeld muss das System eine Reihe von Anforderungen erfüllen. Diese werden im Folgenden beschrieben.

*Schnell.* Um effizient eingesetzt werden zu können, muss die Einrichtezeit auch bei kleinen Losgrößen signifikant kleiner sein als die Zeit für die manuelle Durchführung. Eine schnelle Einrichtung senkt nicht nur die direkten Produktionskosten, sondern reduziert auch die Hemmschwelle, das System für kleinere Losgrößen oder wechselnde Aufgaben einzusetzen. Dadurch wird das System vielseitiger und für eine größere Bandbreite von Anwendungen attraktiv.

*Intuitiv bedienbar.* Durch die intuitive Bedienbarkeit soll das System auch von Mitarbeitern ohne tiefgehende IT- oder Robotik-Kenntnisse genutzt werden können. Dadurch können prozesserfahrene Mitarbeitende ihr domänenspezifisches Wissen direkt in das Automatisierungssystem einbringen, ohne auf die Vermittlung durch externe Inbetriebnehmende angewiesen zu sein.

*Transparent.* Durch eine transparente Kommunikation des Verhaltens kann das Vertrauen in das System gestärkt werden (Kizilcec 2016; Schraagen et al. 2021). Mit dem Vertrauen in KI-Systeme steigt auch deren Akzeptanz (Choung et al. 2023). Mit der Transparenz soll also die Akzeptanz des Systems gefördert werden.

*Kontrolle beim Menschen.* Trotz Automatisierung und KI-Unterstützung soll der Mensch stets die Kontrolle über das System behalten (Sartori & Theodorou 2022). Dies umfasst die Möglichkeit, Eingriffe vorzunehmen, sowie Mechanismen, die sicherstellen, dass Entscheidungen des Systems überprüft und angepasst werden können. Eine solche menschzentrierte Steuerung ist entscheidend, um Vertrauen in das System zu gewährleisten und sicherheitstechnische Anforderungen zu erfüllen.

*Flexibel.* Das System muss sich dynamisch an wechselnde Produktionsbedingungen anpassen können. Dazu gehört die Fähigkeit, auf Variationen in Position und Gestalt von Objekten zu reagieren. Zudem sollte das System kompatibel mit unterschied-lichen Robotermodellen und deren spezifischen Anforderungen sein, um eine breite Einsatzfähigkeit zu gewährleisten.

*Multimodal*. Für eine natürliche und kontextgerechte Interaktion sollen verschiedene Modalitäten wie Sprache, Gestik oder visuelle Hinweise kombiniert werden. Dadurch ist der Mensch nicht nur auf Textbefehle beschränkt, sondern kann seine Umgebung dynamisch zum Erstellen des gewünschten Ablaufes heranziehen.

## 4. Interaktionskonzept

Basierend auf den Anforderungen wurde ein Konzept für die intuitive Programmierung von Robotern entwickelt, welches den natürlichen Erklärprozess zwischen zwei Werkern widerspiegeln soll (Abbildung 1). In einer typischen Einarbeitungssituation wird durch Beschreibungen, Gestik, Rückfragen und Bestätigung miteinander interagiert, um den Arbeitsablauf zu klären und zu überprüfen. Diese Dynamik soll durch die Umsetzung mehrerer Module erreicht werden.

Das zentrale Element zur Programmierung ist die Kommunikation über natürliche Sprache. Nutzende formulieren Handlungsanweisungen oder Arbeitsziele, die vom System interpretiert und in Roboteraktionen übersetzt werden. Durch gezielte Rückfragen des Systems werden Lücken im Ablauf iterativ vervollständigt und Missverständnisse geklärt. Durch Zugriff auf interne Datenbanken, Modelle, Betriebsanleitungen und Stücklisten wird das System darüber hinaus dazu befähigt, Vorschläge für eine Optimierung des Programms zu äußern, die vom Werker beurteilt werden können.

Neben Sprache wird oft auch Gestik zur Vermittlung und Beschreibung von Prozessen genutzt, da sie intuitiv ist und einen zusätzlichen Informationsgehalt vermittelt. Dieser Informationsgehalt soll auch bei der Programmerstellung genutzt werden, indem Anwender auf Bauteile, Werkzeuge oder Vorrichtungen zeigen, bestimmte Positionen spezifizieren oder Bahnen vorgeben können. Ergänzt durch Sprache können diese Objekte und Positionen mit einer semantischen Beschreibung versehen werden, was dem System zusätzlich beim Verständnis der Zielsetzung unterstützt.

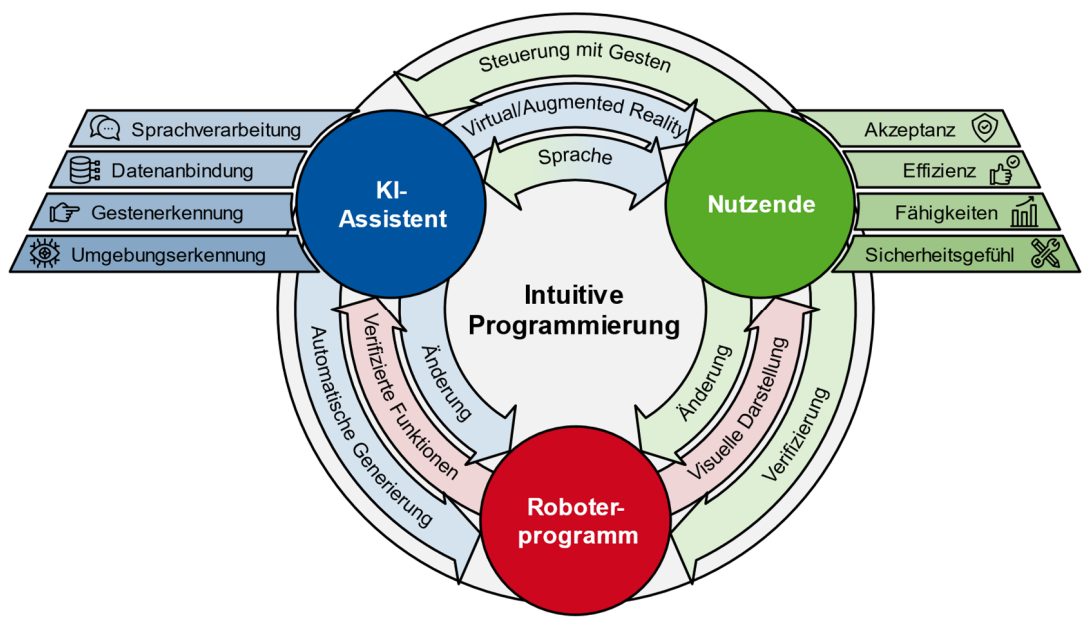

**Abbildung 1:** *Interaktion zwischen den Nutzenden, dem KI-Assistenten und dem erzeugten Programm beim intuitiven Programmieren.*

Über den Dialog hinaus soll den Nutzenden auch ein visuelles Feedback bereitgestellt werden. Eine zentrale Herausforderung besteht darin, Positionen und Bahnen im Raum für den Nutzenden nachvollziehbar zu gestalten. Um dies zu erleichtern, wird eine Augmented-Reality (AR) Anwendung integriert, die die aktuelle Szene durch zusätzliche digitale Informationen ergänzt. Diese Darstellung ermöglicht es den Nutzenden direkt zu sehen und nachzuvollziehen, welche Positionen und Bewegungsbahnen eingelernt wurden. Gleichzeitig werden die entsprechenden Informationen, die aus der sprachlichen Beschreibung abgeleitet wurden, visuell verknüpft dargestellt. Durch diese erweiterte Visualisierung können Nutzende den Programmablauf nicht nur besser verstehen, sondern auch effizienter bearbeiten und anpassen. Dies steigert die Transparenz der Robotersteuerung und vereinfacht die Validierung der programmierten Abläufe erheblich, wodurch die Benutzerfreundlichkeit des Systems weiter erhöht wird.

Mit den validierten Programmen kann der Roboter Aufgaben mehrfach und ohne ständige Überwachung ausführen. Zudem können bereits geprüfte Funktionen in neuen Programmen wiederverwendet werden. Dies reduziert die Notwendigkeit, neuen Code zu schreiben und minimiert Fehlermöglichkeiten.

## 5. Technisches Konzept

Die technische Konzeption des beschriebenen Interaktionskonzepts basiert auf einer Kombination moderner KI- und AR- Technologien. Als zentrale Komponente zur Verarbeitung natürlicher Sprache und zum Generieren des Roboterprogramms soll ein LLM eingesetzt werden. Solche Modelle sind in der Lage, komplexe, mehrschrittige Aufgaben in ihrem Kontext zu verstehen und entsprechend zu strukturieren. Dies umfasst das Identifizieren von Teilaufgaben, das logische Sequenzieren von Arbeitsschritten und die Erstellung kohärenter, auf die Zielanforderungen abgestimmter Programme (Zhou et al. 2024). Darüber hinaus können LLMs natürliche Sprache in standardisierte Datenformate übersetzen, sodass sie durch Robotersysteme weiterverarbeitet werden können. Für das LLM wird eine Dokumentation zu den Basisfunktionen bereitgestellt, die Zweck und Funktionsweise der einzelnen Blöcke erklärt. Um die Wiederverwendbarkeit der automatisch generierten Funktionen zu gewährleisten, wird bei deren Erstellung stets auch eine Dokumentation erzeugt und gespeichert. So kann das LLM bereits validierte Funktionsblöcke wiederverwenden.

Moderne VR- und AR-Brillen bieten die notwendige Hardware für die Hand- und Gestensteuerung sowie für die dreidimensionale Visualisierung im Raum. Neue Wegpunkte können so zum Beispiel durch Tippen mit dem Finger in der Luft gesetzt und dargestellt werden. Das LLM kann durch eine kurze sprachliche Beschreibung des Punktes durch die Anwendenden Attribute ergänzen, die dem Punkt eine Funktion geben und in der Visualisierung entsprechend gekennzeichnet werden.

Für ein grundlegendes Verständnis der Arbeitsumgebung können neuartige CV-Modelle eingesetzt werden. Segment Anything Modelle (SAM) sind in der Lage, Entitäten in einem Bild zu segmentieren und zu identifizieren. In Kombination mit der Gestenerkennung könnten so Objekte im Kontext der Umgebung ausgewählt und zusätzlich semantisch beschrieben werden.

## 6. Diskussion

Das vorgestellte Konzept adressiert zentrale Herausforderungen der modernen Produktionslandschaft wie die steigenden Anforderungen an flexible Produktionsprozesse und die damit einhergehende komplexe Inbetriebnahme-prozesse. LLMs, fortschrittliche CV-Ansätze und multimodale Schnittstellen bieten die Möglichkeit, natürliche Sprache, Gestik und andere kontextbezogene Inputs in die Steuerung von Robotern zu integrieren. Diese Technologien schaffen die Grundlage für eine intuitive und nahtlose Mensch-Maschine-Interaktion, die das Potenzial hat, die Automatisierung grundlegend zu verändern. Trotz dieser Fortschritte bleibt es Gegenstand aktueller Forschung herauszufinden, welche spezifischen Informationen und multimodalen Inputs tatsächlich notwendig sind, um eine Interaktion als „natürlich" wahrzunehmen und die Komplexität und Dynamik realer Arbeitsprozesse abbilden zu können. Dabei sollen besonders Ergonomie, Effektivität und Akzeptanz solcher Systeme im Fokus zukünftiger Untersuchungen stehen. Dazu wird im nächsten Schritt auf dem dargelegten Konzept aufbauend ein Demonstrator entwickelt, mit dem das System anhand verschiedener Use Cases erprobt werden kann.

Mithilfe des entwickelten Systems soll weiter erforscht werden, wie natürliche Interaktionsformen dazu beitragen können, die Automatisierung manueller Tätigkeiten zugänglicher und effizienter zu gestalten. Die tatsächliche Anwendung in der Praxis wird jedoch davon abhängen, wie benutzerfreundlich und robust die entwickelten Lösungen sind, um den Anforderungen industrieller Umgebungen gerecht werden. Eine erfolgreiche Integration erfordert die enge Abstimmung auf die Bedürfnisse der Nutzenden sowie die kontinuierliche Weiterentwicklung von Mensch-Roboter-Interaktionsmodellen.

## 7. Literatur